\documentclass{article}
\usepackage{ijcai20}
\usepackage{times}
\usepackage{epsfig}
\usepackage{graphicx}
\usepackage{amsmath}
\usepackage{amssymb}

\usepackage{color}
\definecolor{citecolor}{RGB}{119,185,0} 
\usepackage[pagebackref=false,breaklinks=true,letterpaper=true,colorlinks,citecolor=citecolor,bookmarks=false]{hyperref}

\usepackage{soul}
\usepackage{url}
\usepackage[utf8]{inputenc}
\usepackage[small]{caption}
\usepackage{multirow}
\usepackage{comment}
\usepackage[dvipsnames,svgnames,x11names]{xcolor}

\usepackage{pifont}
\def\eg{\emph{e.g.}} 
\def\ie{\emph{i.e.}} 
\def\etal{\emph{et~al.}} 

\newlength\savewidth\newcommand\shline{\noalign{\global\savewidth\arrayrulewidth
  \global\arrayrulewidth 1pt}\hline\noalign{\global\arrayrulewidth\savewidth}}

\begin{document}

\title{Unsupervised Scene Adaptation with Memory Regularization \emph{in vivo}}

\author{
  Zhedong Zheng \quad Yi Yang  \\ 
  \affiliations
  Centre for Artificial Intelligence, University of Technology Sydney  \\
  \emails
zhedong.zheng@student.uts.edu.au, yi.yang@uts.edu.au
}

\maketitle

\begin{abstract}
We consider the unsupervised scene adaptation problem of learning from both labeled source data and unlabeled target data. Existing methods focus on minoring the inter-domain gap between the source and target domains. However, the intra-domain knowledge and inherent uncertainty learned by the network are under-explored. In this paper, we propose an orthogonal method, called memory regularization \emph{in vivo} to exploit the intra-domain knowledge and regularize the model training. Specifically, we refer to the segmentation model itself as the memory module, and minor the discrepancy of the two classifiers, \ie, the primary classifier and the auxiliary classifier, to reduce the prediction inconsistency.
Without extra parameters, the proposed method is complementary to most existing domain adaptation methods and could generally improve the performance of existing methods. Albeit simple, we verify the effectiveness of memory regularization on two synthetic-to-real benchmarks: GTA5 $\rightarrow$ Cityscapes and SYNTHIA $\rightarrow$ Cityscapes, yielding $+11.1\%$ and $+11.3\%$ mIoU improvement over the baseline model, respectively. Besides, a similar $+12.0\%$ mIoU improvement is observed on the cross-city benchmark:  Cityscapes $\rightarrow$ Oxford RobotCar.
\end{abstract}

\section{Introduction}
Due to the unaffordable cost of the segmentation annotation, unsupervised scene adaptation is to adapt the learned model to a new domain without extra annotation. In contrast to the conventional segmentation tasks, unsupervised scene adaptation reaches one step closer to the real-world practice. In the real-world scenario, the annotation of the target scene is usually hard to acquire. In contrast, abundant source data is easy to access. To improve the model scalability on the unlabeled target domain, most researchers resort to transfer the common knowledge learned from the source domain to the target domain.

The existing scene adaptation methods typically focus on reducing the discrepancy between the source domain and the target domain. The alignment between the source and target domains could be conducted on different levels, such as pixel level \cite{hoffman2018cycada,wu2018dcan},  feature level \cite{hoffman2018cycada,huang2018domain,yue2019domain,luo2019taking,zhang2019manifold} and semantic level \cite{tsai2018learning,tsai2019domain,wang2019class}. Despite the great success, the brute-force alignment drives the model to learn the domain-agnostic shared features of both domains. 
We consider that this line of methods is sub-optimal in that it ignores the domain-specific feature learning on the target domain, and compromise the final adaptation performance. 

\begin{figure}[t]
\begin{center}
     \includegraphics[width=1\linewidth]{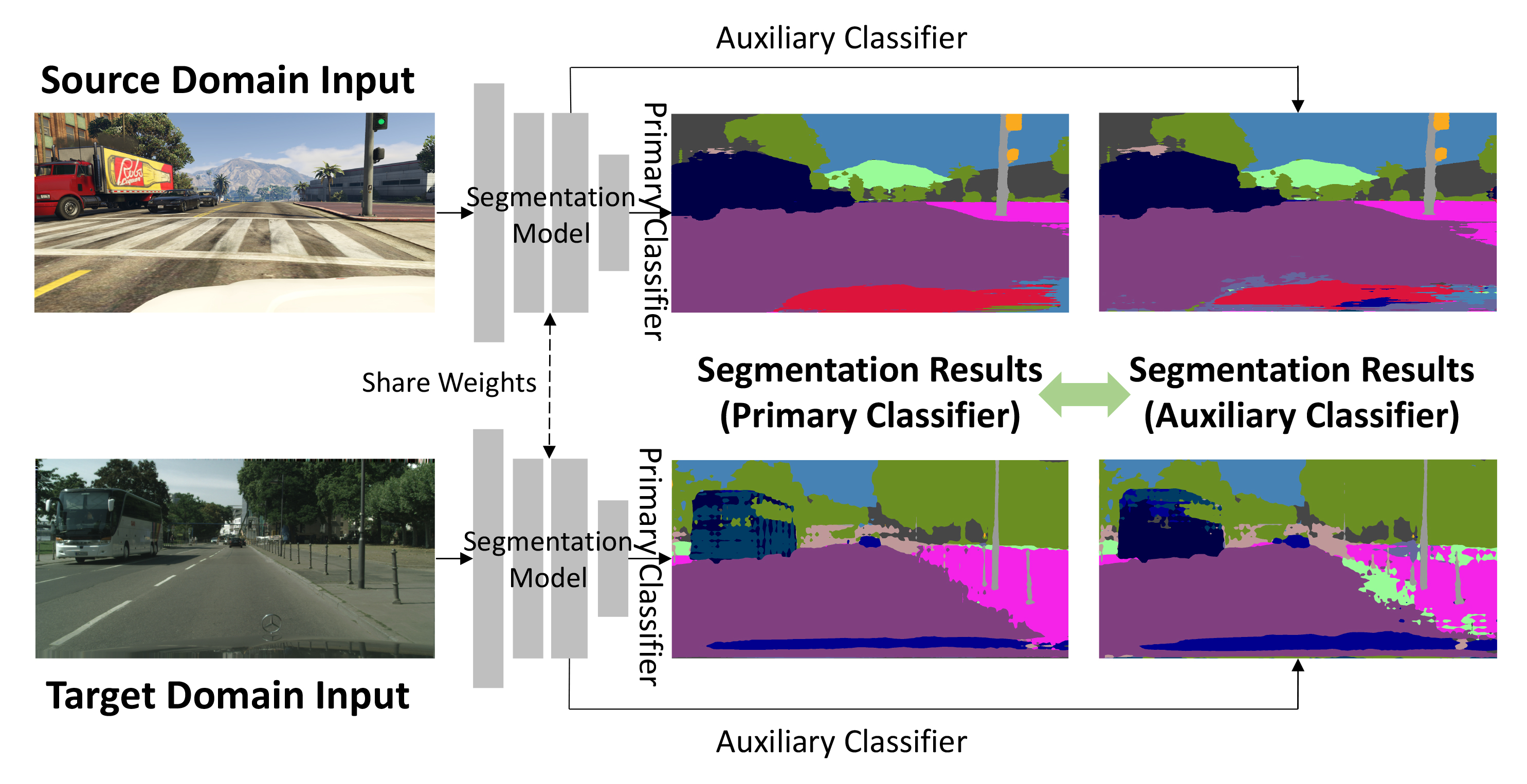}
\end{center} \vspace{-.2in}
      \caption{ We leverage the auxiliary classifier of the widely-used baseline model~\protect\cite{tsai2018learning} to pinpoint the intra-domain uncertainty. While the predictions of the source domain input are relatively consistent, the unlabeled input from the target domain suffers from the uncertain prediction. The model provides different class predictions for the same pixel. It implies that the intra-domain consistency is under-explored, especially in the unlabeled target domain. In contrast to the existing works, which focus on the inter-domain alignment, we focus on one orthogonal direction of mining intra-domain knowledge. }
      \label{fig:consistency}
\end{figure}

\begin{figure*}[t]
\begin{center}
     \includegraphics[width=1\linewidth]{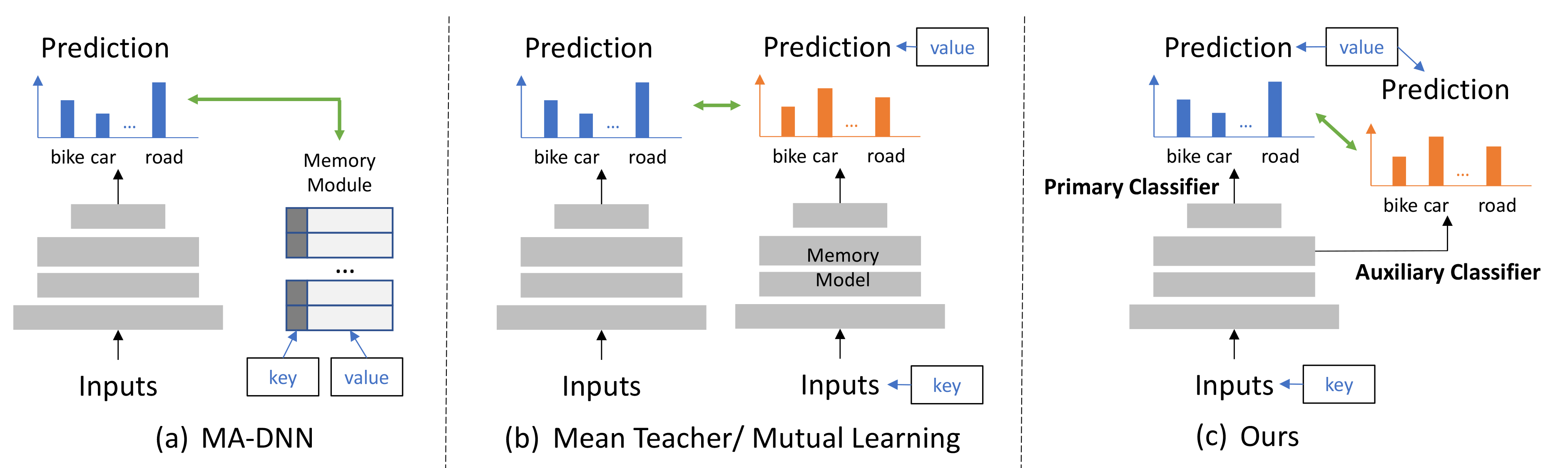}
\end{center}
\vspace{-.1in}
      \caption{ Different Memory-based Methods: (a) MA-DNN~\protect\cite{chen2018semi} applies an extra memory module to save the class prediction while training. (b) Mean teacher~\protect\cite{tarvainen2017mean} and mutual learning~\protect\cite{zhang2018deep} apply one external model to memorize predictions and regularize the training. (c) Different from existing methods, the proposed method does not need extra modules or external models. We leverage the running network itself, as the memory model. Given one input sample as the \emph{key}, we could obtain the two predictions (\emph{values}) from the primary classifier and the auxiliary classifier. 
      }\label{fig:method}
\end{figure*}

Since the domain-specific knowledge is ignored for the target unlabeled data, the regularization by the data itself does not aid in the domain adaptation. To qualitatively verify this, we leverage the auxiliary classifier of the baseline model \cite{tsai2018learning} as a probe to pinpoint the inconsistency. As shown in Fig.~\ref{fig:consistency}, we observe that the model predicts one consistent supervised result of the source labeled data, while the unlabeled target data suffers from the  inconsistency. The predicted result of the primary classifier is different from the auxiliary classifier prediction, especially in the target domain. It implies that the intra-domain consistency has not been learned automatically, when we minor the inter-domain discrepancy.

To effectively exploit the intra-domain knowledge and reduce the target prediction inconsistency, we propose a memory mechanism into the deep neural network training, called memory regularization \emph{in vivo}. 
Different from the previous works focusing on the inter-domain alignment, the proposed method intends to align the different predictions within the same domain to regularize the training. 
As shown in Fig.~\ref{fig:method}(c), we consider the inputs as \emph{key} and the output prediction as the corresponding \emph{value}. In other words, the proposed method deploys the model itself as the \emph{memory module}, which memorizes the historical prediction and learns the key-value projection. Since we have the auxiliary classifier and the primary classifier, we could obtain two \emph{values} for one \emph{key}.  We note that the proposed method is also different from other semi-supervised works deploying the extra memory terms. Since the proposed method does not require additional parameters or modules, we use the term ``\emph{in vivo}'' to differentiate our method from \cite{chen2018semi,tarvainen2017mean,zhang2018deep}; these methods deploy external memory modules.

\textbf{Our contribution} is two-fold: \textbf{(1)} We propose to leverage the memory of model learning to pinpoint the prediction uncertainty and exploit the intra-domain knowledge. This is in contrast to most existing adaption methods focusing on the inter-domain alignment. \textbf{(2)} We formulate the memory regularization \emph{in vivo} as the internal prediction discrepancy between the two classifiers. Different from the existing memory-based models, the proposed method does not need extra parameters, and is compatible with most scene segmentation networks.

\begin{figure*}[t]
\begin{center}
     \includegraphics[width=1\linewidth]{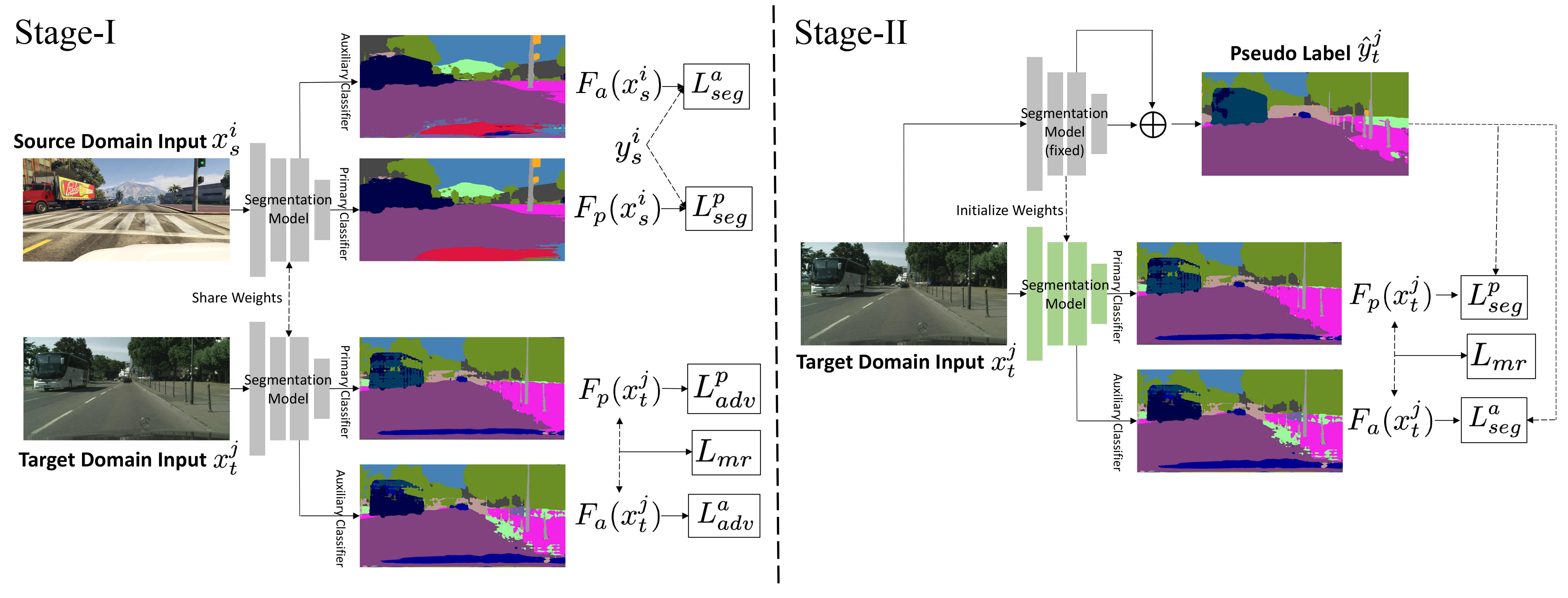}
\end{center}
\vspace{-.15in}
      \caption{Overview of the proposed framework. In the Stage-I, we train the model with the source domain input $x_s^i$ and the target domain input $x_t^i$ to learn the inter-domain and intra-domain knowledge. In the Stage-II, the model focus on the target-domain data and is further fine-tuned with pseudo labels. The proposed memory regularization $L_{mr}$ is applied to regularize the model training in both stages, yielding the performance improvement. 
      }\label{fig:framework}
\end{figure*}

\section{Related Works}
\subsection{Domain Adaptation for Segmentation}
Most existing works typically focus on minoring the domain discrepancy between the source domain and the target domain to learn the shared knowledge. Some pioneering works \cite{hoffman2018cycada,wu2018dcan} apply the image generator to  transfer the source data to the style of the target data, and intend to reduce the low-level visual appearance difference. Similarly, Yu \etal \cite{yue2019domain} and Wu \etal \cite{wu2019ace} generate the training images of different styles to learn the domain-agnostic feature. Adversarial loss is also widely studied. Tsai \etal \cite{tsai2018learning,tsai2019domain} apply the adversarial losses to different network layers to enforce the domain alignment. Luo \etal \cite{luo2019taking} leverage the attention mechanism and the class-aware adversarial loss to further improve the performance. 
Besides, some works also focus on mining the target domain knowledge, which is close to our work. Zou \etal \cite{zou2018unsupervised,zou2019confidence} leverage the confident pseudo labels to further fine-tune the model on the target domain, yielding a competitive benchmark. Recently, Shen \etal \cite{shen2019regularizing} propose to  utilize the discriminator to find the confident pseudo label. Different from the pseudo label based methods, the proposed method focuses on target domain knowledge by mining the intrinsic uncertainty of the model learning on the unlabeled target-domain data. We note that the proposed method is orthogonal to the existing methods, including the inter-domain alignment \cite{tsai2018learning,tsai2019domain,luo2019taking} and self-training with pseudo labels \cite{zou2018unsupervised,zou2019confidence}. In Section \ref{sec:ablation}, we show the proposed method can be integrated with other domain adaption methods to further improve the performance. 

\subsection{Memory-based Learning}
As one of the early works, Weston \etal \cite{weston2014memory} propose to use external memory module to store the long-term memory. In this way, the model could reason with the related experience more effectively. 
Chen \etal \cite{chen2018semi} further apply the memory to the semi-supervised learning to learn from the unlabeled data. In this work, we argue that the teacher model, which is applied in many frameworks, also could be viewed as one kind of external memory terms. Because the teacher model distills the knowledge of the original setting, and memorizes the key concepts to the final prediction \cite{hinton2015distilling}. For instance, one of the early work, called temporal ensemble \cite{laine2016temporal}, uses the historical models to regularize the running model, yielding the competitive performance. The training sample could be viewed as the key, and the historical models are the memory model to find the corresponding value for the key. Since the historical models memorize the experience from the previous training samples, the temporal ensemble could provide stable and relatively accurate predictions of the unlabeled data. Except for \cite{laine2016temporal}, there are different kinds of external memory models. 
Mean Teacher \cite{tarvainen2017mean} leverages the weight moving average model as the memory model to regularize the training. Further, French \etal \cite{french2017self} extend Mean Teacher for visual domain adaptation. Zhang \etal \cite{zhang2018deep} propose mutual learning, which learns the knowledge from multiple student models. 

Different from existing memory-based methods \cite{chen2018semi,tarvainen2017mean,zhang2018deep}, the proposed method leverages the memory of the model itself to regularize the running model. The proposed memory regularization does not introduce extra parameters and external modules. (see Fig.~\ref{fig:method})

\section{Method}

\subsection{Algorithm Overview}
\noindent\textbf{Formulation.} We denote the images from the source domain and the target domain as $X_s = \{x_s^i\}_{i=1}^M$ and $X_t = \{x_t^j\}_{j=1}^N$, where $M,N$ are the number of the source images and target images. Every source domain data in $X_s$ is annotated with corresponding ground-truth segmentation maps $Y_s = \{y_s^i\}_{i=1}^M$. 
Given one unlabeled target domain image $x_t^j$, we intend to learn a function to project the image to the segmentation map $y_t^j$. 
Following the practice in \cite{tsai2018learning,luo2019taking}, we adopt the modified DeepLabv2 as our baseline model, which contains one backbone model and two classifiers, \ie, the primary classifier $C_p$ and the auxiliary classifier $C_a$. To simplify, we denote the two functions $F_p$ and $F_a$ as the segmentation functions, where $F_p$ projects the image to the prediction of the primary classifier, and $F_a$ maps the input to the prediction of the auxiliary classifier.

\noindent\textbf{Overview.} As shown in Fig.~\ref{fig:framework}, the proposed method has two training stages, \ie, Stage-I and Stage-II, to progressively transfer the learned knowledge from the labeled source data to the unlabeled target data. In the Stage-I, we follow the conventional domain adaptation methods to minor the inter-domain discrepancy between the source domain and the target domain. When training, we regularize the model by adding the memory regularization. The memory regularization helps to minor the intra-domain inconsistency, yielding the performance improvement. 
In the Stage-II, we leverage the trained model to predict the label for the unlabeled target data. Then the model is further fine-tuned on the target domain. With the help of pseudo labels, the model could focus on learning domain-specific knowledge on the target domain. 
The pseudo labels inevitably contain noise, and the memory regularization in Stage-II could prevent the model from overfitting to the noise in pseudo labels. 
Next we introduce different objectives for the model adaptation in detail. We divide the losses into two classes: (1) Domain-agnostic learning to learn the shared inter-domain features from the source domain; (2) Domain-specific learning to learn the intra-domain knowledge, especially the features for the target domain.

\subsection{Domain-agnostic learning}
\noindent\textbf{Segmentation loss.} 
First, we leverage the annotated source-domain data to learn the source-domain knowledge. The segmentation loss is widely applied, and could be formulated as the pixel-wise cross-entropy loss: 
\begin{align}
    L_{seg}^p &= -\sum_{h=1}^{H}\sum_{w=1}^{W}\sum_{c=1}^{C}{y_s^i\log( F_p(x_s^i))}, \\
    L_{seg}^a &= -\sum_{h=1}^{H}\sum_{w=1}^{W}\sum_{c=1}^{C}{y_s^i\log( F_a(x_s^i))},
\end{align}
where the first loss is for the primary prediction, and the second objective is for the auxiliary prediction. $H$ and $W$ denote the height and the width of the input image, and $C$ is the number of segmentation classes. 

\noindent\textbf{Adversarial loss.} Segmentation loss only focuses on the source domain. 
We demand one objective to minor the discrepancy of the target domain and the source domain, and hope that the model could transfer the source-domain knowledge to the target domain. We, therefore, introduce the adversarial loss \cite{tsai2018learning} to minor the discrepancy of the source domain and the target domain. The adversarial loss is applied to both predictions of the primary classifier and the auxiliary classifier:
\begin{align}
    L_{adv}^p &= \mathbb{E}[ \log(D_p(F_p(x_s^i))) + \log(1-D_p(F_p(x_t^j))) ], \\
    L_{adv}^a &= \mathbb{E}[ \log(D_a(F_a(x_s^i))) + \log(1-D_a(F_a(x_t^j))) ],
\end{align}
where $D$ denotes the discriminator. In this work, we deploy two different discriminators, \ie, $D_p$ and $D_a$, for the primary prediction and the auxiliary prediction, respectively. The discriminator is to find out whether the target prediction $F(x_t)$ is close to the source prediction $F(x_s)$ in the semantic space. By optimizing the adversarial loss, we force the model to bridge the inter-domain gap on the semantic level.
 
\subsection{Domain-specific learning}
However, the segmentation loss and the adversarial loss do not solve the intra-domain inconsistency, especially in the target domain. In the Stage-I, we consider leveraging the uncertainty in the target domain and propose the memory regularization \emph{in vivo} to enforce the consistency. In the Stage-II, we further utilize the memory to regularize the training and prevent the model overfitting to the noisy pseudo labels.

\noindent\textbf{Memory regularization.}
In this paper, we argue that the model itself could be viewed as one kind of memory module, in that the model memorizes the historical experience. Without introducing extra parameters or external modules, we enforce the model to learn from itself. In particular, we view the input image as the \emph{key}, and the model as the \emph{memory module}.
Given the input image (\emph{key}), the model could generate the \emph{value} by simply feeding forward the \emph{key}. We could obtain two \emph{values} by the primary classifier and the auxiliary classifier, respectively. To minor the uncertainty of the model learning on the target domain, we hope that the two \emph{values} of the same \emph{key} could be as close to each other as possible, so we deploy the KL-divergence loss:
\begin{equation}
\begin{aligned}
    L_{mr} = &-\sum_{h=1}^{H}\sum_{w=1}^{W}\sum_{c=1}^{C}{F_a(x_t^i)\log( F_p(x_t^i))} \\
    &-\sum_{h=1}^{H}\sum_{w=1}^{W}\sum_{c=1}^{C}{F_p(x_t^i)\log( F_a(x_t^i))}.
\end{aligned}
\end{equation}
We only apply the memory regularization loss on the target domain $X_t$ and ask the mapping functions $F_a$ and $F_p$ to generate a consistent prediction on the unlabeled target data. 

\noindent Discussion. 
1. \textbf{What is the advantage of the memory regularization?} 
By using the memory regularization, we enable the model to learn the intra-domain knowledge on the unlabeled target data with an explicit and complementary objective.
As discussed in the \cite{tarvainen2017mean,chen2018semi}, we could not ensure that the memory always provides a right class prediction for the unlabeled data. The memory mechanism is more likely to act as a teacher model, providing the class distribution based on the historical experience. 
2.\textbf{Will the auxiliary classifier hurt the primary classifier?} As shown in many semi-supervised methods~\cite{zhang2018deep,tarvainen2017mean}, the bad-student model also could provide essential information for the top-student models. Our experiment also verifies that the sub-optimal auxiliary classifier could help the primary classifier learning, and vice versa (see Section \ref{sec:ablation}). 

\noindent\textbf{Self-training with pseudo labels.}
In the Stage-II, we do not use the source data anymore. The model is fine-tuned on the unlabeled target data and mine the target domain knowledge. Following the self-training policy in \cite{zou2018unsupervised,zou2019confidence}, we retrain the model with the pseudo label $\hat{y}_t^j$. The pseudo label combines the output of $F_p(x_t^j)$ and $F_a(x_t^j)$ from the trained model in the Stage-I. In particular, we set the $\hat{y}_t^j = \operatorname*{arg\,max}(F_p(x_t^j) + 0.5F_a(x_t^j))$. 
The pseudo segmentation loss could be formulated as:
\begin{align}
    L_{pseg}^p &= -\sum_{h=1}^{H}\sum_{w=1}^{W}\sum_{c=1}^{C}{\hat{y}_t^j\log( F_p(x_t^j))}, \\
    L_{pseg}^a &= -\sum_{h=1}^{H}\sum_{w=1}^{W}\sum_{c=1}^{C}{\hat{y}_t^j\log( F_a(x_t^j))}.
\end{align}
We apply the pixel-wise cross-entropy loss as the supervised segmentation loss. Since most pseudo labels are correct, the model still could learn from the noisy labels. In Section \ref{sec:ablation}, we show the self-training with pseudo labels further boosts the performance on the target domain despite  the noise in pseudo labels.

\noindent Discussion. 
\textbf{What is the advantage of the memory regularization in the Stage-II?} In fact, we treat the pseudo labels as the supervised annotations in the Stage-II. However, the pseudo labels contain the noise and may mislead the model to overfit the noise. The proposed memory regularization in the Stage-II works as a smoothing term, which enforces the consistency in the model prediction, rather than focusing on fitting the pseudo label extremely. 

\subsection{Optimization}
We integrate the above-mentioned losses. The total loss of the Stage-I and Stage-II training could be formulated as:
\begin{align}
L_{S1}(F_a,F_p,D_a,D_p) &= L_{seg} + L_{adv} + \lambda_{mr}L_{mr}, \\
L_{S2}(F_a,F_p) &= L_{pseg} + \lambda_{mr}L_{mr},
\end{align}
where $\lambda_{mr}$ is the weight for the memory regularization. We follow the setting in PSPNet \cite{zhao2017pyramid} to set $0.5$ for segmentation losses on the auxiliary classifier. $L_{seg} = L_{seg}^p + 0.5L_{seg}^a$, $L_{pseg} = L_{pseg}^p + 0.5L_{pseg}^a$. For adversarial losses, we follow the setting in \cite{tsai2018learning,luo2019taking}, and select small weights for adversarial loss terms $L_{adv} = 0.001L_{adv}^p + 0.0002L_{adv}^a$. Besides, we fix the weight of memory regularization as $\lambda_{mr}=0.1$ for all experiments.

\subsection{Implementation Details}
\textbf{Network Architectures.} We deploy the widely-used Deeplab-v2 \cite{chen2017deeplab} as the baseline model, which adopts the ResNet-101 \cite{he2016deep} as the backbone model. Since the auxiliary classifier has been widely adopted in the scene segmentation frameworks, such as PSPNet \cite{zhao2017pyramid} and modified DeepLab \cite{tsai2018learning,luo2019taking}, for fair comparison, we also applied the auxiliary classifier in our baseline model as well as the final full model. We also insert the dropout layer before the classifier layer, and the dropout rate is $0.1$.
Besides, we follow the PatchGAN \cite{isola2017image} and deploy the multi-scale discriminator model.

\noindent\textbf{Implementation Details.}  The input image is resized to $1280 \times 640$, and we randomly crop $1024 \times 512$ for training. We deploy the SGD optimizer with the batch size $2$ for the segmentation model, and the initial learning rate is set to $0.0002$. The optimizer of the discriminator is Adam and the learning rate is set to $0.0001$. Following \cite{zhao2017pyramid,zhang2019dual}, both segmentation model and discriminator deploy the ploy learning rate decay by multiplying the factor $(1-\frac{iter}{total-iter})^{0.9}$. We set the total iteration as $100k$ iteration and adopt the early-stop policy.
The model is first trained without the memory regularization for $10k$ to avoid the initial prediction noise, and then we add the memory regularization to the model training. For Stage-I, we train the model with $25k$ iterations. We further fine-tune the model in the Stage-II for $25k$ iterations. 
We also adopt the class balance policy in the \cite{zou2018unsupervised} to increase the weight of the rare class, and the small-scale objects.  When inference, we combine the outputs of both classifiers $\hat{y}_t^j = \operatorname*{arg\,max}(F_p(x_t^j) + 0.5F_a(x_t^j))$. Our implementation is based on Pytorch. We will release our code for reproducibility. 

\begin{table}[tbp]
\small
\begin{center}
{
\setlength{\tabcolsep}{10pt}
\begin{tabular}{l|c|c}
\shline
Method & without $L_{mr}$ & with $L_{mr}$ \\
\hline
Auxiliary Classifier & 40.04  &  44.45\\
Primary Classifier  & 43.11  &  45.29 \\
\hline
Ours (Stage-I) & 42.73 & 45.46 \\
\shline
\end{tabular}}
\end{center}
\vspace{-2.5mm}
\caption{Ablation study of the memory regularization on both classifiers, \ie, the auxiliary classifier and the primary classifier, in the Stage-I training. The result suggests that the memory regularization helps both classifiers, especially the auxiliary classifier. The final results of the full model combine the results of both classifiers, and therefore improve the performance further.}\label{table:classifier}
\end{table}

\begin{table}[tbp]
\small
\begin{center}
{
\setlength{\tabcolsep}{7pt}
\begin{tabular}{l|c|c|c|c}
\shline
Method & $L_{seg}$ & $L_{adv}$ & $L_{mr}$ & mIoU \\
\hline
Without Adaptation & $\checkmark$ & & & 37.23\\
Adversarial Alignment  & $\checkmark$ & $\checkmark$ & & 42.73 \\
Memory Regularization & $\checkmark$ &  & $\checkmark$ & 43.75 \\
Ours (Stage-I)  & $\checkmark$ & $\checkmark$ & $\checkmark$ & 45.46 \\
\shline
\end{tabular}}
\end{center}
\vspace{-2.5mm}
\caption{Ablation study of different losses in the Stage-I training. We gradually add the adversarial loss $L_{adv}$ and the memory regularization $L_{mr}$ into consideration.} \label{table:stage1}
\end{table}

\section{Experiment}
\subsection{Dataset and Evaluation Metric}
We mainly evaluate the proposed method on the two unsupervised scene adaption settings, \ie, GTA5 \cite{richter2016playing} $\rightarrow$ Cityscapes \cite{cordts2016cityscapes} and SYNTHIA \cite{ros2016synthia} $\rightarrow$ Cityscapes \cite{cordts2016cityscapes}. Both source datasets, \ie, GTA5 and SYNTHIA, are the synthetic datasets. GTA5 contains $24,966$ training images, while SYNTHIA has $9,400$ images for training. The target dataset, Cityscapes, is collected in the realistic scenario, including $2,975$ unlabeled training images. Besides, we also evaluate the proposed method on the cross-city benchmark: Cityscapes \cite{cordts2016cityscapes} $\rightarrow$ Oxford RobotCar \cite{RobotCarDatasetIJRR}. We follow the setting in \cite{tsai2019domain} and evaluate the model on the Cityscapes validation set/ RobotCar validation set. 
For the evaluation metric, we report the mean Intersection over Union (mIoU), averaged over all classes.

\subsection{Ablation Studies} \label{sec:ablation}
\noindent\textbf{Effect of the memory regularization.} 
To investigate how the memory helps both classifiers, we report the results of the single classifier in Table~\ref{table:classifier}. The observation suggests two points: First, memory regularization helps both classifier learning and improves the performance of both classifiers, especially the auxiliary classifier. Second, the accuracy of the primary classifier prediction does not decrease due to the relatively poor results of the auxiliary classifier. The primary classifier also increases by $2.18\%$ mIoU. It verifies that the proposed memory regularization helps to reduce the inconsistency and mine intra-domain knowledge.
Furthermore, we report the results of the full model after Stage-I training, which combines the predictions of both classifiers. The full model arrives $45.46\%$ mIoU accuracy, which is slightly higher than the prediction accuracy of the primary classifier. It also indicates that the predictions of the auxiliary classifier and primary classifier are complementary.

\begin{table}[tbp]
\small
\begin{center}
{
\setlength{\tabcolsep}{14pt}
\begin{tabular}{l|c|c|c}
\shline
Method & $L_{pseg}$ & $L_{mr}$  & mIoU \\
\hline
Ours (Stage-I) &  & &  45.46 \\
Pseudo Label & $\checkmark$ & &  47.90\\
Ours (Stage-II) & $\checkmark$ & $\checkmark$ &  48.31 \\
\shline
\end{tabular}}
\end{center}
\vspace{-2.5mm}
\caption{Ablation study of different losses in the Stage-II training. The result suggests that the memory regularization could prevent the model from overfitting to the noise in the pseudo labels.} \label{table:stage2}
\end{table}

\begin{table*}[!t]
	\centering
	\resizebox{\linewidth}{!}{
	\begin{tabular}{c|c|ccccccccccccccccccc|c}
		\shline
		Method & Backbone & Road & SW & Build & Wall & Fence & Pole & TL & TS & Veg. & Terrain & Sky & PR & Rider & Car & Truck & Bus & Train & Motor & Bike & mIoU\\
		\shline
		Source & \multirow{2}{0.1\linewidth}{\centering{DRN-26}} & 42.7 & 26.3 & 51.7 & 5.5 & 6.8 & 13.8 & 23.6 & 6.9  & 75.5 & 11.5 & 36.8 & 49.3 & 0.9 & 46.7 & 3.4 & 5.0 & 0.0 & 5.0 & 1.4  & 21.7\\
		CyCADA~\cite{hoffman2018cycada} & & 79.1 & 33.1 & 77.9 & 23.4 & 17.3 & 32.1 & 33.3 & 31.8 & 81.5 & 26.7 & 69.0 & 62.8 & 14.7 & 74.5 & 20.9 & 25.6 & 6.9 & 18.8 & 20.4 & 39.5\\
		\hline
		Source & \multirow{2}{0.1\linewidth}{\centering{DRN-105}} & 36.4 & 14.2 & 67.4 & 16.4 & 12.0 & 20.1 & 8.7 & 0.7 & 69.8 & 13.3 & 56.9 & 37.0 & 0.4 & 53.6 & 10.6 & 3.2 & 0.2 & 0.9 & 0.0 & 22.2\\
		MCD~\cite{saito2018maximum} & & 90.3 & 31.0 & 78.5 & 19.7 & 17.3 & 28.6 & 30.9 & 16.1 & 83.7 & 30.0 & 69.1 & 58.5 & 19.6 & 81.5 & 23.8 & 30.0 & 5.7 & 25.7 & 14.3 & 39.7\\
		\hline
		Source & \multirow{6}{0.1\linewidth}{\centering{DeepLabv2}} & 75.8 & 16.8 & 77.2 & 12.5 & 21.0 & 25.5 & 30.1 & 20.1 & 81.3 & 24.6 & 70.3 & 53.8 & 26.4 & 49.9 & 17.2 & 25.9 & 6.5 & 25.3 & 36.0 & 36.6\\
		AdaptSegNet~\cite{tsai2018learning} & & 86.5 & 36.0 & 79.9& 23.4 & 23.3 & 23.9 & 35.2 & 14.8 & 83.4 & 33.3 & 75.6 & 58.5 & 27.6 & 73.7 & 32.5 & 35.4 & 3.9 & 30.1 & 28.1 & 42.4\\ 
		SIBAN \cite{luo2019significance} & & 88.5 & 35.4 & 79.5 & 26.3 & 24.3 & 28.5 & 32.5 & 18.3 & 81.2 & 40.0 & 76.5 & 58.1 & 25.8 & 82.6 & 30.3 & 34.4 & 3.4 & 21.6 & 21.5 & 42.6 \\
		CLAN~\cite{luo2019taking} & & 87.0 & 27.1 & 79.6 & 27.3 & 23.3 & 28.3 & 35.5 & 24.2 & 83.6 & 27.4 & 74.2 & 58.6 & 28.0 & 76.2 & 33.1 & 36.7 & 6.7 & 31.9 & 31.4 & 43.2 \\
		APODA~\cite{yang2020adversarial} & & 85.6 & 32.8 & 79.0 & 29.5 & 25.5 & 26.8 & 34.6 & 19.9 & 83.7 & \textbf{40.6} & 77.9 & 59.2 & 28.3 & 84.6 & 34.6 & 49.2 & 8.0 & \textbf{32.6} & 39.6 & 45.9 \\
		PatchAlign~\cite{tsai2019domain} & & \textbf{92.3} & 51.9 & 82.1 & 29.2 & 25.1 & 24.5 & 33.8 & 33.0 & 82.4 & 32.8 & 82.2 & 58.6 & 27.2 & 84.3 & 33.4 & \textbf{46.3} & 2.2 & 29.5 & 32.3 & 46.5 \\
		\hline
		AdvEnt~\cite{vu2019advent} & DeepLabv2 & 89.4 & 33.1 & 81.0 & 26.6 & \textbf{26.8} & 27.2 & 33.5 & 24.7 & 83.9 & 36.7 & 78.8 & 58.7 & 30.5 & 84.8 & \textbf{38.5} & 44.5 & 1.7 & 31.6 & 32.4 & 45.5 \\
		\hline
		Source & \multirow{2}{0.1\linewidth}{\centering{DeepLabv2}} &  - & - & - & - & - & - & - & - & -& - & - & - & - & - & - & - & - & - & - & 29.2\\
		FCAN~\cite{zhang2018fully} &  & - & - & - & - & - & - & - & - & -& - & - & - & - & - & - & - & - & - & - & 46.6 \\
		\hline
		Source & \multirow{3}{0.1\linewidth}{\centering{DeepLabv2}} & 71.3 & 19.2 & 69.1 & 18.4 & 10.0 & 35.7 & 27.3 &  6.8 & 79.6 & 24.8 & 72.1 & 57.6 & 19.5 & 55.5 & 15.5 & 15.1 & 11.7 & 21.1 & 12.0 & 33.8\\
		CBST \cite{zou2018unsupervised} & & 91.8 & 53.5 & 80.5 & 32.7 & 21.0 & 34.0 & 28.9 & 20.4 & 83.9 & 34.2 & 80.9 & 53.1 & 24.0 & 82.7 & 30.3 & 35.9 & 16.0 & 25.9 & \textbf{42.8} & 45.9\\
		MRKLD \cite{zou2019confidence} & & 91.0 & \textbf{55.4} & 80.0 & 33.7 & 21.4 & \textbf{37.3} & 32.9 & 24.5 & 85.0 & 34.1 & 80.8 & 57.7 & 24.6 & 84.1 & 27.8 & 30.1 & \textbf{26.9} & 26.0 & 42.3 & 47.1\\
		\hline
		Source & \multirow{3}{0.1\linewidth}{\centering{DeepLabv2}} & 51.1 & 18.3 & 75.8 & 18.8 & 16.8 & 34.7 & 36.3 & 27.2 & 80.0 & 23.3 & 64.9 & 59.2 & 19.3 & 74.6 & 26.7 & 13.8 & 0.1 & 32.4 & 34.0 & 37.2\\
		Our (Stage-I)  & & 89.1 & 23.9 & 82.2 & 19.5 & 20.1 & 33.5 & 42.2 & 39.1 & \textbf{85.3} & 33.7 & 76.4 & 60.2 & 33.7 & \textbf{86.0} & 36.1 & 43.3 & 5.9 & 22.8 & 30.8 & 45.5 \\
		Our (Stage-II) & & 90.5 & 35.0 & \textbf{84.6} & \textbf{34.3} & 24.0 & 36.8 & \textbf{44.1} & \textbf{42.7} & 84.5 & 33.6 & \textbf{82.5} & \textbf{63.1} & \textbf{34.4} & 85.8 & 32.9 & 38.2 & 2.0 & 27.1 & 41.8 & \textbf{48.3} \\
		\shline
	\end{tabular}
	}
	\vspace{-2.5mm}
	\caption{Quantitative results on GTA5 $\rightarrow$ Cityscapes. We present pre-class IoU and mIoU. The best accuracy in every column is in \textbf{bold}.}
	\label{table:gtacity}
\end{table*}

\begin{table*}[!t]
	\centering
	\resizebox{\linewidth}{!}{
	\begin{tabular}{c|c|cccccccccccccccc|c|c}
		\shline
		Method & Backbone & Road & SW & Build & Wall* & Fence* & Pole* & TL & TS & Veg. & Sky & PR & Rider & Car & Bus & Motor & Bike & mIoU* & mIoU\\
		\shline
		Source & \multirow{2}{0.1\linewidth}{\centering{DRN-105}} & 14.9 & 11.4 & 58.7 & 1.9 & 0.0 & 24.1 & 1.2 & 6.0 & 68.8 & 76.0 & 54.3 & 7.1 & 34.2 & 15.0 & 0.8 & 0.0 & 26.8 & 23.4\\
		MCD \cite{saito2018maximum} & & 84.8 & \textbf{43.6} & 79.0 & 3.9 & 0.2 & 29.1 & 7.2 & 5.5 & 83.8 & 83.1 & 51.0 & 11.7 & 79.9 & 27.2 & 6.2 & 0.0 & 43.5 & 37.3 \\
		\hline
		Source & \multirow{6}{0.1\linewidth}{\centering{DeepLabv2}} & 55.6 & 23.8 & 74.6 & $-$ & $-$ & $-$ & 6.1 & 12.1 & 74.8 & 79.0 & 55.3 & 19.1 & 39.6 & 23.3 & 13.7 & 25.0 & 38.6 & $-$ \\
		SIBAN~\cite{luo2019significance} & & 82.5 & 24.0 & 79.4 & $-$ & $-$ & $-$ & 16.5 & 12.7 & 79.2 & 82.8 & 58.3 & 18.0 & 79.3 & 25.3 & 17.6 & 25.9 & 46.3 & $-$ \\
    	PatchAlign~\cite{tsai2019domain} & & 82.4 & 38.0 & 78.6 & 8.7 & 0.6 & 26.0 & 3.9 & 11.1 & 75.5 & 84.6 & 53.5 & 21.6 & 71.4 & 32.6 & 19.3 & 31.7 & 46.5 & 40.0 \\
		AdaptSegNet~\cite{tsai2018learning} & & 84.3 & 42.7 & 77.5 & $-$ & $-$ & $-$ & 4.7 & 7.0 & 77.9 & 82.5 & 54.3 & 21.0 & 72.3 & 32.2 & 18.9 & 32.3 & 46.7 & $-$ \\
		CLAN~\cite{luo2019taking} & & 81.3 & 37.0 & 80.1 & $-$ & $-$ & $-$ & 16.1 & 13.7 & 78.2 & 81.5 & 53.4 & 21.2 & 73.0 & 32.9 & 22.6 & 30.7 & 47.8 & $-$  \\
		APODA~\cite{yang2020adversarial} & & \textbf{86.4} & 41.3 & 79.3 & $-$ & $-$ & $-$ & 22.6 & 17.3 & 80.3 & 81.6 & 56.9 & 21.0 & 84.1 & \textbf{49.1} & \textbf{24.6} & 45.7 & 53.1 & $-$  \\
		\hline
		AdvEnt \cite{vu2019advent} & DeepLabv2 & 85.6 & 42.2 & 79.7 & 8.7 & 0.4 & 25.9 & 5.4 & 8.1 & 80.4 & \textbf{84.1} & 57.9 & 23.8 & 73.3 & 36.4 & 14.2 & 33.0 & 48.0 & 41.2 \\
		\hline
		Source & \multirow{3}{0.1\linewidth}{\centering{DeepLabv2}} & 64.3 & 21.3 & 73.1 & 2.4 & 1.1 & 31.4 & 7.0 & 27.7 & 63.1 & 67.6 & 42.2 & 19.9 & 73.1 & 15.3 & 10.5 & 38.9 & 40.3 & 34.9 \\
		CBST \cite{zou2018unsupervised} & & 68.0 & 29.9 & 76.3 & \textbf{10.8} & 1.4 & 33.9 & 22.8 & 29.5 & 77.6 & 78.3 & 60.6 & 28.3 & 81.6 & 23.5 & 18.8 & 39.8 & 48.9 & 42.6 \\
		MRKLD \cite{zou2019confidence} & & 67.7 & 32.2 & 73.9 & 10.7 & \textbf{1.6} & \textbf{37.4} & 22.2 & \textbf{31.2} & 80.8 & 80.5 & 60.8 & \textbf{29.1} & 82.8 & 25.0 & 19.4 & 45.3 & 50.1 & 43.8 \\
		\hline
		Source & \multirow{3}{0.1\linewidth}{\centering{DeepLabv2}} & 44.0 & 19.3 & 70.9 & 8.7 & 0.8 & 28.2 & 16.1 & 16.7 & 79.8 & 81.4 & 57.8 & 19.2 & 46.9 & 17.2 & 12.0 & 43.8 & 40.4 & 35.2 \\
		Ours (Stage-I) & & 82.0 & 36.5 & 80.4 & 4.2 & 0.4 & 33.7 & 18.0 & 13.4 & 81.1 & 80.8 & 61.3 & 21.7 & 84.4 & 32.4 & 14.8 & 45.7 & 50.2 & 43.2 \\
		Ours (Stage-II) & & 83.1 & 38.2 & \textbf{81.7} & 9.3 & 1.0 & 35.1 & \textbf{30.3} & 19.9 & \textbf{82.0} & 80.1 & \textbf{62.8} & 21.1 & \textbf{84.4} & 37.8 & 24.5 & \textbf{53.3} & \textbf{53.8} & \textbf{46.5} 
		\\
		\shline
	\end{tabular}
	}
	\vspace{-2.5mm}
	\caption{Quantitative results on SYNTHIA $\rightarrow$ Cityscapes. We present pre-class IoU, mIoU and mIoU*. mIoU and mIoU* are averaged over 16 and 13 categories, respectively. The best accuracy in every column is in \textbf{bold}.}
	\label{table:syncity}
\end{table*}

\noindent\textbf{Effect of different losses in Stage-I.}
As shown in Table \ref{table:stage1}, the full model could improve the performance from $37.23\%$ to $45.46\%$ mIoU. When only using the adversarial loss $L_{adv}$, the model equals to the widely-used domain adaptation method \cite{tsai2018learning}. We note that the model only using the memory regularization $L_{mr}$ also achieves significant improvement comparing to the baseline model without adaption. We speculate that the memory regularization helps to mine the target domain knowledge, yielding the better performance on the target domain. After combining all three loss terms, the full model arrives $45.46\%$ mIoU on Cityscapes.

\noindent\textbf{Effect of different losses in Stage-II.} If we only deploy the pseudo segmentation loss $L_{pseg}$, the model equals to several previous self-training methods \cite{zou2018unsupervised,zou2019confidence}. However, this line of previous methods usually demands a well-designed threshold for the label confidence. In contrast, we do not introduce any threshold, but apply the memory regularization to prevent overfitting toward noisy labels. As shown in Table \ref{table:stage2}, the full model with memory regularization arrives $48.31\%$ mIoU accuracy, which is higher than the result of the model only trained on the pseudo labels. It verifies that the proposed memory regularization also helps the model learning from noisy labels.

\begin{figure}[t]
\begin{center}
     \includegraphics[width=1\linewidth]{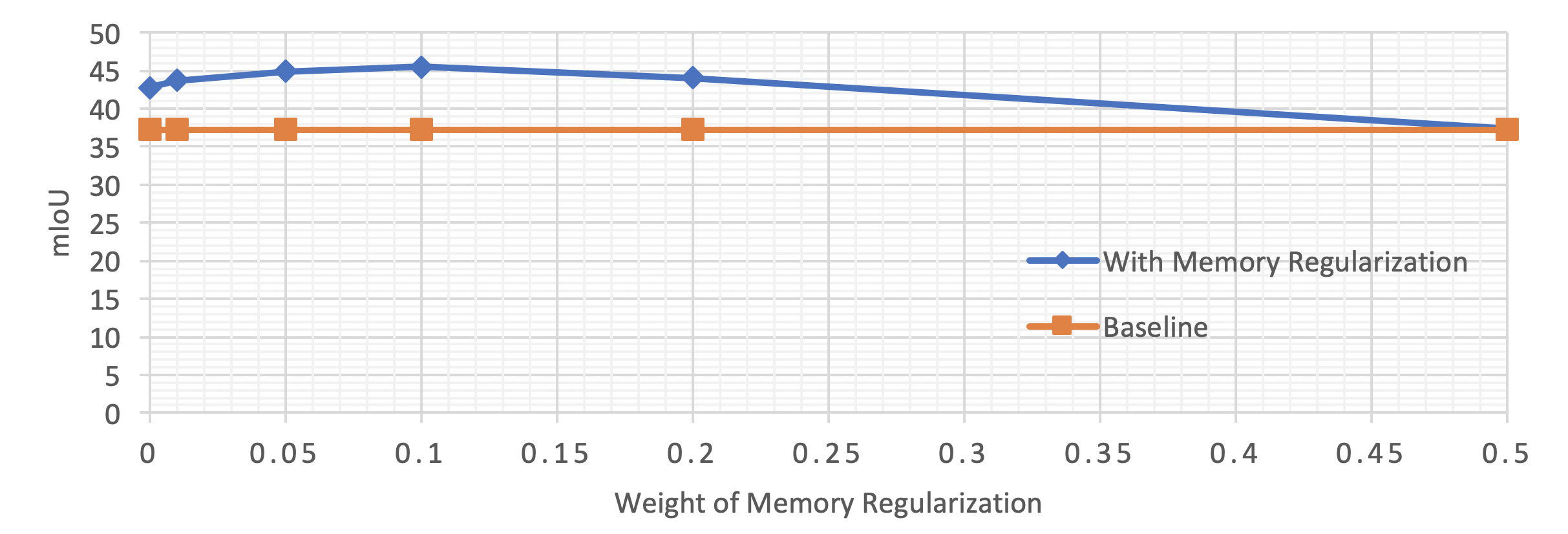}
\end{center}
\vspace{-.2in}
      \caption{Sensitivity of mIoU to the hyper-parameter $\lambda_{mr}$ on Cityscapes.}
      \label{fig:hyperparameter}
\end{figure}

\noindent\textbf{Hyperparameter Analysis.} In this work, we introduce $\lambda_{mr}$ as the weight of the memory regularization. As shown in Fig.~\ref{fig:hyperparameter}, we evaluate different weight values $\{0,0.01,0.05,0.1,0.2,0.5\}$. We observe that the model is robust to the value of $\lambda_{mr}$. However, when the value $\lambda_{mr}$ is too large or small, the model may mislead to overfitting or underfitting the consistency. Therefore, without loss of generality, we use $\lambda_{mr} = 0.1$ for all experiments.

\subsection{Comparisons with state-of-the-art methods}
\textbf{Synthetic-to-real.} We compare the proposed method with different domain adaptation methods on GTA5 $\rightarrow$ Cityscapes (See Table~\ref{table:gtacity}). For a fair comparison, we mainly show the results based on the same backbone, \ie, DeepLabv2. The proposed method has achieved $48.3\%$ mIoU, which is higher than the competitive methods, \eg, pixel-level alignment \cite{hoffman2018cycada}, semantic level alignment \cite{tsai2018learning}, as well as the self-training methods, \ie, \cite{zou2018unsupervised,zou2019confidence}. 
Compared with the strong source-only model, the proposed method yields $+11.1\%$ improvement. Besides, we observe a similar result on SYNTHIA $\rightarrow$ Cityscapes (see Table~\ref{table:syncity}). The proposed method arrives $53.8\%$ mIoU* and $46.5\%$ mIoU, which is also competitive to other methods. We obtain $+11.3\%$ improvement in mIoU accuracy over the baseline.

\noindent\textbf{Cross-city.} We also evaluate the proposed method on adapting the model between different cities. The two real datasets, \ie, Cityscapes and Oxford RobotCar, are different from collection locations as well as weather conditions. Cityscapes is collected in the sunny days when Oxford RobotCar contains rainy scenarios. As shown in Table \ref{table:oxford}, the proposed method also achieves competitive results, \ie, $73.9\%$ mIoU.    

\begin{table} [!t]
	\centering
    \resizebox{\linewidth}{!}{
	\begin{tabular}{l|ccccccccc|c}
		\shline
		Method & \rotatebox{90}{road} & \rotatebox{90}{sidewalk} & \rotatebox{90}{building} & \rotatebox{90}{light} & \rotatebox{90}{sign} & \rotatebox{90}{sky} & \rotatebox{90}{person} & \rotatebox{90}{automobile} & \rotatebox{90}{two-wheel} & mIoU\\
		
		\hline
		
		Without Adaptation & 79.2 & 49.3 & 73.1 & 55.6 & 37.3 & 36.1 & 54.0 & 81.3 & 49.7 & 61.9 \\
		AdaptSegNet~\cite{tsai2018learning}  & 95.1 & 64.0 & 75.7 & 61.3 & 35.5 & 63.9 & 58.1 & 84.6 & 57.0 & 69.5 \\
		PatchAlign~\cite{tsai2019domain} & 94.4 & 63.5 & 82.0 & 61.3 & 36.0 & 76.4 & 61.0 & 86.5 & 58.6 & 72.0 \\
		\hline
		Ours (Stage-I) & \textbf{95.9} & \textbf{73.5} & 86.2 & 69.3 & 31.9 & 87.3 & 57.9 & 88.8 & \textbf{61.5} & 72.5 \\
		Ours (Stage-II)  & 95.1 & 72.5 & \textbf{87.0} & \textbf{72.2} & \textbf{37.4} & \textbf{87.9} & \textbf{63.4} & \textbf{90.5} & 58.9 & \textbf{73.9} \\
		\shline
	\end{tabular}}
	\vspace{-2.5mm}
	\caption{
		Quantitative results on the cross-city benchmark: Cityscapes $\rightarrow$ Oxford RobotCar.
	}
		\label{table:oxford}
\end{table}

\section{Conclusion}
We propose a memory regularization method for unsupervised scene adaption. Our model leverages the intra-domain knowledge and reduces the uncertainty of model learning. Without introducing any extra parameters or external modules, we deploy the model itself as the memory module to regularize the training. Albeit simple, the proposed method is complementary to previous works and achieves competitive results on two synthetic-to-real benchmarks, \ie, GTA5 $\rightarrow$ Cityscapes and SYNTHIA $\rightarrow$ Cityscapes, and one cross-city benchmark, \ie, Cityscapes $\rightarrow$ Oxford RobotCar. 

{\footnotesize
\bibliographystyle{named}
\bibliography{egbib}
}

\end{document}